\documentclass[runningheads]{llncs}
\usepackage{graphicx}
\usepackage{xcolor}
\usepackage{amsmath}
\usepackage{amssymb}
\usepackage{multirow}
\usepackage{booktabs}
\usepackage{hyperref}
\usepackage[T1]{fontenc}

\begin{document}
%
\title{Topological SLAM in colonoscopies leveraging deep features and topological priors }
\titlerunning{Topological SLAM in colonoscopies}
%
\author{Javier Morlana, Juan D. Tardós and José M. M. Montiel}

%
\authorrunning{Javier Morlana et al.}
%
\institute{i3A, University of Zaragoza, Spain \\
\email{\{jmorlana,tardos,josemari\}@unizar.es}}
\maketitle              
\begin{abstract}

We introduce ColonSLAM, a system that combines classical multiple-map metric SLAM with deep features and topological priors to create topological maps of the whole colon. The SLAM pipeline by itself is able to create disconnected individual metric submaps representing locations from short video subsections of the colon, but is not able to merge covisible submaps due to deformations and the limited performance of the SIFT descriptor in the medical domain. ColonSLAM is guided by topological priors and combines a deep localization network trained to distinguish if two images come from the same place or not and the soft verification of a transformer-based matching network, being able to relate far-in-time submaps during an exploration, grouping them in nodes imaging the same colon place, building more complex maps than any other approach in the literature. We demonstrate our approach in the Endomapper dataset, showing its potential for producing maps of the whole colon in real human explorations. Code and models are available at: \href{https://github.com/endomapper/ColonSLAM}{github.com/endomapper/ColonSLAM}

\end{abstract}
\section{Introduction}

\label{section:intro}

The interest for medical Computer Vision has been growing in the last decades, with many works being able to successfully apply classical and modern techniques to the medical domain. In this context, Simultaneous Localization And Mapping (SLAM) is a line of research that has been receiving huge attention due to its broad spectrum of possible applications such as medical robotics and navigation assistance. SLAM systems aim to localize a camera while building a map of an unexplored environment. Two kinds of representations can be obtained by SLAM algorithms. Metric SLAM estimates a 6DoF camera trajectory and a geometric 3D point cloud, while topological SLAM obtains a graph whose nodes represent places that can be connected by covisibility or traversability. 

We are interested in the colonoscopy domain, a field of medicine in which technology still has little presence. Typically, practitioners manoeuvre through the colon anatomy based on prior knowledge and experience, visualizing the raw endoscopy images in the screen without any other information input. However, metric SLAM struggles in colonoscopies due to illumination changes that hinder keyframe registration, and dynamic elements and deformations that violate the rigidity constraint. The result are small and disconnected 3D submaps, quite different from the long maps obtained in out-of-the-body scenes. 

We propose ColonSLAM, a topological SLAM system for the colonoscopy domain, where nodes are groups of small metric submaps imaging the same colon place.  We build on top of a recent metric SLAM that builds small submaps using classical image features, and we continuosly find relationships between far-in-time submaps, leveraging deep global visual place recognition descriptors, transformer-based matching techniques and topological connectivity priors. Our contributions in this work are threefold:

\begin{itemize}
    \item We present ColonSLAM, the first metric-topological SLAM system able to map the whole colon creating a graph that codes the procedure complexity.
    \item We propose a novel visual place recognition network $\mathbb{L}$, able to identify covisible images to build a topological map from submaps obtained by metric SLAM.
    \item We perform an evaluation in real human colonoscopy data, showing our ability to build complex maps to cover the entire colon exploration.
\end{itemize}


\section{Related Work}

\label{sec:related}

\noindent \textbf{Metric SLAM} solutions already work well in natural scenes, being able to map unknown environments through feature-based approaches \cite{mur2015orb,campos2021orb}, which employ geometric bundle adjustment, or direct methods such as \cite{engel2014lsd,engel2017direct}, optimizing errors in the photometric space.  Nowadays, there is a growing interest in bringing SLAM to the medical domain. Mahmoud et al \cite{Mahmoud2019} applies ORB-SLAM \cite{mur2015orb} to laparoscopy, SAGE-SLAM \cite{liu2022sage} integrates learned depth and features to reconstruct endonasal surgery scenes, and RNN-SLAM \cite{ma2021rnnslam} combines DSO \cite{engel2017direct} with learned depth to create dense reconstructions of the colon. The recent approach CudaSIFT-SLAM \cite{elvira2024cudasift} builds on the ORB-SLAM3 multi-mapping system \cite{campos2021orb} replacing ORB features by CudaSIFT \cite{cudasift2007}, building metric multi-maps in human colon in real-time. It produces small disjoint 3D maps, where covisibility between the keyframes in each map is guaranteed as every keyframe goes through several stages of filtering: matching, geometric verification, 3D triangulation and geometric bundle adjustment. Multi-maps are key for robustly dealing with tracking losses due to occlusions, deformation and motion blur prevalent in colonoscopy.

RNN-SLAM and CudaSIFT-SLAM are currently the top performers in colonoscopic SLAM, but they are unable to relate far-in-time submaps representing the same place. We build on the output of CudaSIFT-SLAM to obtain meaningful topological maps by establishing relationships between their disjoint submaps.

\noindent \textbf{Topological SLAM} avoids the geometry estimation and focuses on aggregating covisible images by their appearance, leveraging on visual place recognition (VPR) methods. These algorithms can be better suited for the medical domain, where metric SLAM tends to fail due to deformations or occlusions. Classical methods \cite{cummins2008fab,angeli2008incremental,galvez2011real} converted local features such as SIFT \cite{lowe2004sift} or ORB \cite{rublee2011orb} into a Bag-of-Words representation, finding the most similar images, further verified by geometry in order to close a loop between nodes. Recently, ColonMapper \cite{morlana2023colonmapper} leveraged the Bayesian filtering proposed in \cite{angeli2008fast,angeli2008incremental} with global deep features for VPR to build topological maps with a trivial two-node connectivity which links each node with its anterior and posterior in time neighbours. Despite its simplicity, ColonMapper is able to map the whole colon, and remarkably, the map was reused for topological localization two weeks afterwards, in a second colonoscopy of the same patient. While ColonMapper builds the map and afterwards localizes, our ColonSLAM performs a proper topological SLAM, simultaneously localizing and updating the map in the processing of each new incoming submap. 

Our proposal is also close to recent works building topological graphs with the help of deep learning \cite{chaplot2020neural,nagarajan2020ego,savinov2018semi}. They build a graph using retrieval networks as in \cite{morlana2023colonmapper}, but tailoring it as means to an end, focusing on robot navigation or affordances learning. Differently from them, we focus on building the graph that defines the topological map, as creating meaningful representations is not straightforward in the medical domain. Colonoscopy images, in particular, are a challenging task for visual recognition algorithms due to their weak texture and the visual similarity of different regions.  

\noindent \textbf{Neural Networks for Visual Place Recognition} are also closely related to our work. The works of \cite{morlana2021self,ma2021colon10k,morlana2023colonmapper} brought popular image retrieval networks \cite{radenovic2018fine,arandjelovic2016netvlad} to the colonoscopy domain, and a similar approach was followed for topological graphs in out-of-the-body scenes in \cite{chaplot2020neural,nagarajan2020ego,savinov2018semi}. ColonMapper \cite{morlana2023colonmapper}, particularly, employed an image retrieval network trained by means of a margin loss and used it to for trivial topological map creation and its posterior localization. We empirically found that this training objective is not discriminative enough to build non-trivial maps with high precision. Our work also leverages on the transformer-based matcher LightGlue \cite{lindenberger2023lightglue} to establish relationships between consecutive nodes, while ColonMapper tried a similar strategy with LoFTR \cite{sun2021loftr}.

\section{ColonSLAM}


\subsection{Node building}
\label{subsec:creation}

We create a topological map $G = (N,E)$ composed by nodes $N$ and edges $E$. Each node represents a \textit{place}, a distinctive section of the colon, while edges link traversable places connected in space. ColonMapper \cite{morlana2023colonmapper} assumed a simplistic graph of consecutive places connected with its two closest neighbours computed from a global descriptor similarity and a matching verification with LoFTR. In contrast, we propose to sequentially build a full-fledged topological map which captures the complex covisibility and traversability among the metric submaps.


\begin{figure}[t]
    \centering
    \includegraphics[width=0.90\columnwidth]{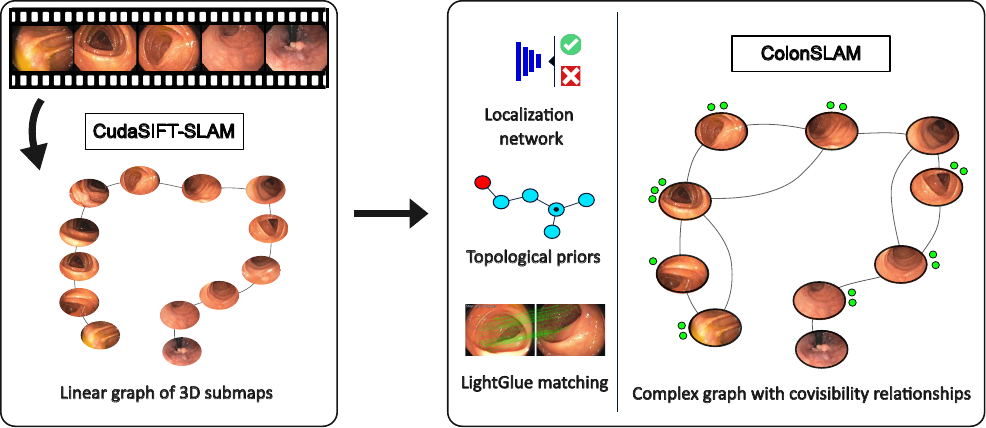}
    \caption{\textbf{ColonSLAM}. From a linear graph of metric submaps, ColonSLAM is able to obtain a topological graph with rich connections by leveraging a novel localization network, topological priors and LightGlue matching.}
    \label{fig:overview}
\end{figure} 

The starting point of the topological map in ColonSLAM is a set of metric 3D disjoint \textit{submaps} obtained by CudaSIFT-SLAM with their linear connectivity, i.e. each submap is connected only with its anterior and posterior submaps (see Fig.\,\ref{fig:overview}). They are composed of several keyframes (distinctive images). For each keyframe, we extract a global descriptor $\textbf{d} \in \mathbb{R}^{D}$ by means of a localization network $\mathbb{L}$ (Sec. \ref{subsec:network}). The submaps obtained are usually small, typically 5 seconds lifespan and 15 keyframes.
Some of the maps are taken from the same colon location, but CudaSIFT-SLAM was not able to merge them together. As our objective is to build rich graphs for the colon, we use a similar formulation as \cite{nagarajan2020ego}, considering our node as a colon region that encompasses several submaps observing that particular region. For example, if CudaSIFT-SLAM reconstructed the submaps $s_{20}, s_{22}$ and $s_{23}$ for the cecum area, our graph would represent the cecum as a node $n \in N$ with submaps $\left\{ s_{20}, s_{22}, s_{23}  \right\}$.  We discard \textit{in-between} frames, chunks of video between submaps that were not included in any CudaSIFT-SLAM 3D model, as they are generally noisy observations containing unmappable frames, i.e. blurry, occluded or covered by fluids.



\subsection{Localization Network}
\label{subsec:network}

\begin{figure}[t]
    \centering
    \includegraphics[width=0.75\columnwidth]{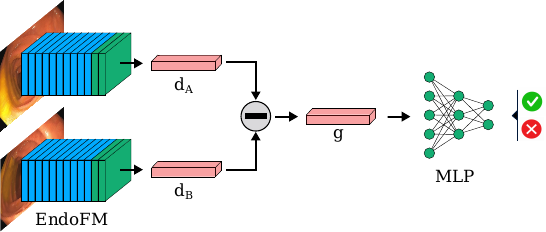}
    \caption{\textbf{ Localization network $\mathbb{L}$}. It obtains a $sim$ score, deciding if two images come from the same place. The backbone green blocks and the MLP are fine-tuned.}
    \label{fig:network}
\end{figure}

Our localization network $\mathbb{L}$ predicts if two images come from the same place or not, and we use it to determine if the incoming \textit{submap} is already included in the map. The network  is composed of a backbone and a 5-layer MLP. The backbone is initialized from the endoscopy foundational model EndoFM \cite{wang2023foundation}, which, for an image $I$ extracts a global descriptor $\textbf{d} \in \mathbb{R}^{768}$. To decide if two images $I_{A}, I_{B}$ come from the same place, we subtract their descriptors $g = d_{A} - d_{B}$ and feed $g$ to the MLP followed by a softmax, predicting a similarity score $sim$ that allows to decide if they come from the same place, as can be observed in Fig. \ref{fig:network}. We fine-tune the last two layers of the backbone and the MLP using a cross-entropy objective. Training details are explained in Sec. \ref{subsec:implementation}.

\subsection{Topological Simultaneous Localization and Mapping}
\label{subsec:topological}

ColonSLAM receives a linear topological graph formed by all submaps from CudaSIFT-SLAM. The main idea of ColonSLAM is to identify which submaps represent the same colon location, merging them into the same node. This observation capability builds traversability links between distant nodes, resulting in a richer graph than the linear one. Previous work \cite{morlana2023colonmapper} followed a Bayesian approach for localizing a new exploration of a patient against a trivial map of this patient built in a previous exploration. We empirically found that using a Bayesian approach to simultaneously map and localize decreases the performance of topological SLAM. For this reason, we opted for a simpler yet effective approach that allows us to leverage on topological priors without explicitly modelling the localization probability. We demonstrate that these priors are helpful for all methods, specially in combination with the discriminating power of our localization network $\mathbb{L}$ and the matching capabilities of LightGlue \cite{lindenberger2023lightglue}. 

\noindent \textbf{Topological priors.} Our topological SLAM starts with the first \textit{submap} $s_{0}$, that initializes the first node $N_{0}$. For every submap $s_{i}$, we extract a descriptor $d_{k}$ with $\mathbb{L}$ for every keyframe $I_{k} \in s_{i}$ . Here we leverage on colonoscopy priors, which are linear by nature. In a typical colonoscopy, the practitioner reaches the cecum as fast as possible, and then, performs a slow exploration in the withdrawal stage. Occasionally, the camera moves back and forth, i.e. when the endoscope is obstructed or the practitioner is exploring carefully a particular area. In any case, the camera has to observe close-by areas to the current location before reaching further places. It is physically impossible to go from the cecum to the transverse without going through the ascending colon first. For this reason, we establish our \textit{topological connectivity prior} as a search space $\omega$ where a covisible node can be looked for. The search space $\omega$ is defined as a window of nodes at a distance equal or smaller than $m$ nodes from the previous position $S_{t-1}$.

\noindent \textbf{Selecting a localization $S_{t}$.} We compute the score $sim$ of $I_{k} \in s_{i}$ against every node $n \in \omega$. We compare $d_k$ against all the image descriptors in each $s_{j} \in n$ with our localization network $\mathbb{L}$. The score for each submap $s_{j}$ is equal to the average of the top-3 ranked images from $s_{j}$, while the score for the node $n$ is the highest score among its submaps. Besides, for every keyframe $I_{k}$ from $s_{i}$ we store in $l_{sim}$ the node $n_{j}$ with the highest score and its value. We define the $score_{\mathbb{L}}$ as the median value of the scores for the node $n_{j}$ that with higher occurences in $l_{sim}$.

We are interested in determining if the incoming submap $s_i$ belongs to a node $n_j$ in the graph or not, in order to add $s_i$ to $n_j$ or create a new node $n_{new}$ connected to the previous position $S_{t-1}$. We have two ways of triggering a localization: a) LightGlue finds $m_{LG} > th_{LG}$ matches between any image in $s_i$ and any image from any node $n_j \in \omega$ (Eq. \ref{eq:a}) or b) $score_{\mathbb{L}} > th_{sim}$ (Eq. \ref{eq:b}): 

\begin{eqnarray}
    m_{LG} & > & th_{LG} \label{eq:a} \\ 
    score_{\mathbb{L}} & > & th_{sim} \label{eq:b} 
\end{eqnarray}

The reasoning behind bypassing LightGlue in condition b) is that, despite its matching abilities and precision, LightGlue is not able to deal with all the challenges in colonoscopies, failing when images are far from each other. $\mathbb{L}$ is able to reliably find some of this cases, so we chose to complement the two methods, looking for higher recall values while keeping an acceptable precision ($\sim90\%$). If a localization is accepted in $n_{j}$, the current position $S_{t}$ is set to $n_{j}$, otherwise is set to $n_{new}$, adding a traversability link with previous position $S_{t-1}$.

\section{Experiments}
\label{sec:experiments}

\subsection{Implementation details} 
\label{subsec:implementation}

\textbf{Localization network $\mathbb{L}$ training.} We train our localization network $\mathbb{L}$ with the Endomapper \cite{azagra2023endomapper} training data proposed in \cite{morlana2023colonmapper}. We use the already labelled data to extract samples. Labels were obtained in \cite{morlana2023colonmapper} using COLMAP \cite{schonberger2016structure} and manual labelling. It includes positive examples from COLMAP and some hard positives manually labelled, besides per-cluster covisibility labelling that allows extracting negative pairs from the same sequence. For our cross-entropy loss, we train with pairs query-positive and query-negative, trying to predict if the images are similar or dissimilar, respectively. We get one random positive sample for each query from the positives pool, while we always provide the hardest negative coming from the same sequence as the query, based on the global descriptor distance. \cite{nagarajan2020ego} trained with concatenated vectors, while we found crucial for our network's convergence to subtract them before passing the result to the MLP. We fine-tune the last two layers of EndoFM \cite{wang2023foundation} and the MLP, freezing the rest of the network. Convergence is achieved after 4 epochs based on the cross-entropy loss in the validation set, using 10k queries per epoch and re-mining negatives every 2500 queries. Our training framework is based on \cite{berton2022deep}.

\noindent\textbf{Other details.} We use off-the-shelf LightGlue \cite{lindenberger2023lightglue}, reducing the SuperPoint detection threshold and disabling early stoppers from LightGlue in order to get the most reliable matches. The matching acceptance threshold is $th_{LG} = 100$.

\subsection{Evaluation on the Endomapper dataset}

\begin{table}[t]
\resizebox{0.95\columnwidth}{!}{%
    \centering
    \begin{tabular}{lccccccc}
        \toprule
        \multirow{2}{*}{Method} & \multicolumn{2}{c}{Seq\_027} & \multicolumn{2}{c}{Seq\_035}  & \multicolumn{2}{c}{Average} & \multirow{2}{*}{Runtime} \\ \cmidrule(lr){2-3} \cmidrule(lr){4-5} \cmidrule(lr){6-7}
                               &      Precision   &     Recall             &     Precision   &     Recall             &      Precision   &     Recall              \\
        \midrule
        Morlana21 \cite{morlana2021self}  	                                     & 0.83 & 0.51 & 0.88 & 0.49 & 0.85 & 0.50 & 38 s \\
        + Topologic prior                 	                                                  & 0.88  & 0.50  & \underline{0.95}  & 0.47 & 0.91 & 0.48 & 36 s \\
        \midrule
        R50-NV-H \cite{morlana2023colonmapper} 	                                 & 0.64 & 0.42 & \textbf{0.97} & 0.37 & 0.80 & 0.39 & 50 s \\
        + Topologic prior             	                                                      & 0.78  & 0.45  & \textbf{0.97}  & 0.33 & 0.87 & 0.39 & 39 s \\
        \midrule
        LightGlue \cite{lindenberger2023lightglue}  	        & \textbf{1.0}  & 0.45 & 0.91 & 0.33 & \underline{0.95} &  0.39 & $\sim$56 min \\
        + Topologic prior     	                                                & \textbf{1.0}  & 0.45  & 0.94  & 0.32 & \textbf{0.97} & 0.38 & $\sim$11 min \\
        \midrule
        $\mathbb{L}$ (ours)  	                                                 & 0.87  & \underline{0.64}  & 0.76  & \underline{0.68} & 0.81 & \underline{0.66} & $\sim$1 min \\
        + Topologic prior                                                                   & \underline{0.96}  & 0.61  & 0.92  & 0.67 & 0.94 & 0.64 & 50 s \\
        + LightGlue	                                                       & 0.94  & \textbf{0.70}  & 0.87  & \textbf{0.70} & 0.90 & \textbf{0.70} & $\sim$25 min \\

        \bottomrule
    \end{tabular}
    }
    \caption{Precision and Recall results. Bold: best. Underlined: second best.}
    \label{table:PR}
\end{table}

We selected two sequences of the Endomapper dataset as our ground truth. We chose the same sequences as ColonMapper (Seq\_027 and Seq\_035), the closest work to ours, easing the comparison. Labeling was done following the text footage available in the Endomapper dataset, created by the doctor during the exploration. We first process them with CudaSIFT-SLAM, obtaining a set of \textit{submaps} $\in \{s_0,\ldots,s_n \}$. We manually labelled which submaps are covisible, that is, should belong to the same node. Two nodes are covisible if they observe the same location. We labelled both medium-covisible relationships and long-term covisibility, i.e. a polyp seen both in the entry and the withdrawal phase. Additionally, \textit{submaps} are labelled chronologically: we know if the incoming \textit{submap} should be localized against previous nodes or if it should create a new node.

We show precision and recall values in Table \ref{table:PR}. We compare against related methods to our work: Morlana21 \cite{morlana2021self} and R50-NV-H (from ColonMapper) \cite{morlana2023colonmapper}, two image retrieval networks trained for the colonoscopy domain, and LightGlue \cite{lindenberger2023lightglue}, a state-of-the-art network in image matching, with enough matching power to establish correspondences between close nodes in colonoscopies. ColonMapper also proposed a localization algorithm where mapping was not considered, so its application here is not straightforward. Instead, we evaluate the network proposed in their work. Besides, we provide an ablation study of the three main elements of our pipeline: the localization network $\mathbb{L}$, the addition of the topological prior and LightGlue matching. Precision and recall are defined as:

\begin{equation}
  \mathbf{P} =\frac{TP}{TP + FP}\;\; , \;\; \mathbf{R} =\frac{TP}{TP + FN}
  \label{eq:eq_PR}
\end{equation}

$TP$ are true positives, correctly localized \textit{submaps}. A localization for \textit{submap} $s_{i}$ for node $n_{j}$ is deemed correct if the majority of \textit{submaps} in $n_{j}$ were labelled as positives with $s_{i}$. $FP$ are false positives, wrongly localized \textit{submaps}. $FN$ are false negatives, \textit{submaps} that should be localized but instead started a new node.

We evaluate the performance of the different methods and the benefits of the topological prior. For retrieval networks (Morlana21 \cite{morlana2021self}, R50-NV-H \cite{morlana2023colonmapper} and $\mathbb{L}$), we accept a localization only if Eq. \ref{eq:b} is fulfilled. We apply a different threshold for each of the networks as the score distribution given by each network is different. To allow a fair comparison between them, we tuned the best threshold for every network in terms of precision-recall performance. For Morlana21 \cite{morlana2021self}, $th_{sim} = 0.85$; for ColonMapper \cite{morlana2023colonmapper}, $th_{sim} = 0.65$ and for $\mathbb{L}$, $th_{sim}=0.95$. For LightGlue \cite{lindenberger2023lightglue}, we compare first, medium and last image in $s_{i}$ against the first, medium and last image of all nodes, as comparing all is too expensive. If any comparison fulfills Eq. \ref{eq:a}, we stop the computation and accept the localization. 

Approaches without the topological prior search along the whole graph, while when the topological prior is added, the search is only allowed in the window $\omega$, with $m = 5$. For our full approach ($\mathbb{L}$ + Topological prior + LightGlue), we accept a localization if Eq. \ref{eq:a} or \ref{eq:b} are fulfilled as explained in Sec. \ref{subsec:topological}. All approaches improve their precision significantly when the topological prior is applied, specially for our network $\mathbb{L}$, that receives a great boost in precision while getting the highest recall. Reducing the search space using the topological graph information is helpful for image retrieval networks, as they are not confused by similar frames coming from far regions. The effect in LightGlue is minimal, as it is only able to match close-by images, but it reduces computation time by 5x while maintaining the performance. Our network $\mathbb{L}$, in combination with the topological prior, is able to compete with LightGlue precision while getting an improvement of $+70\%$ in recall an being several orders of magnitude faster. Despite this, we aim to find at most connections as possible (high recall) while having a reasonable precision. When complementing our approach with LightGlue, we finally obtain a precision of $90\%$ with a recall of $70\%$.

\begin{figure}[t]
    \centering
    \includegraphics[width=\columnwidth]{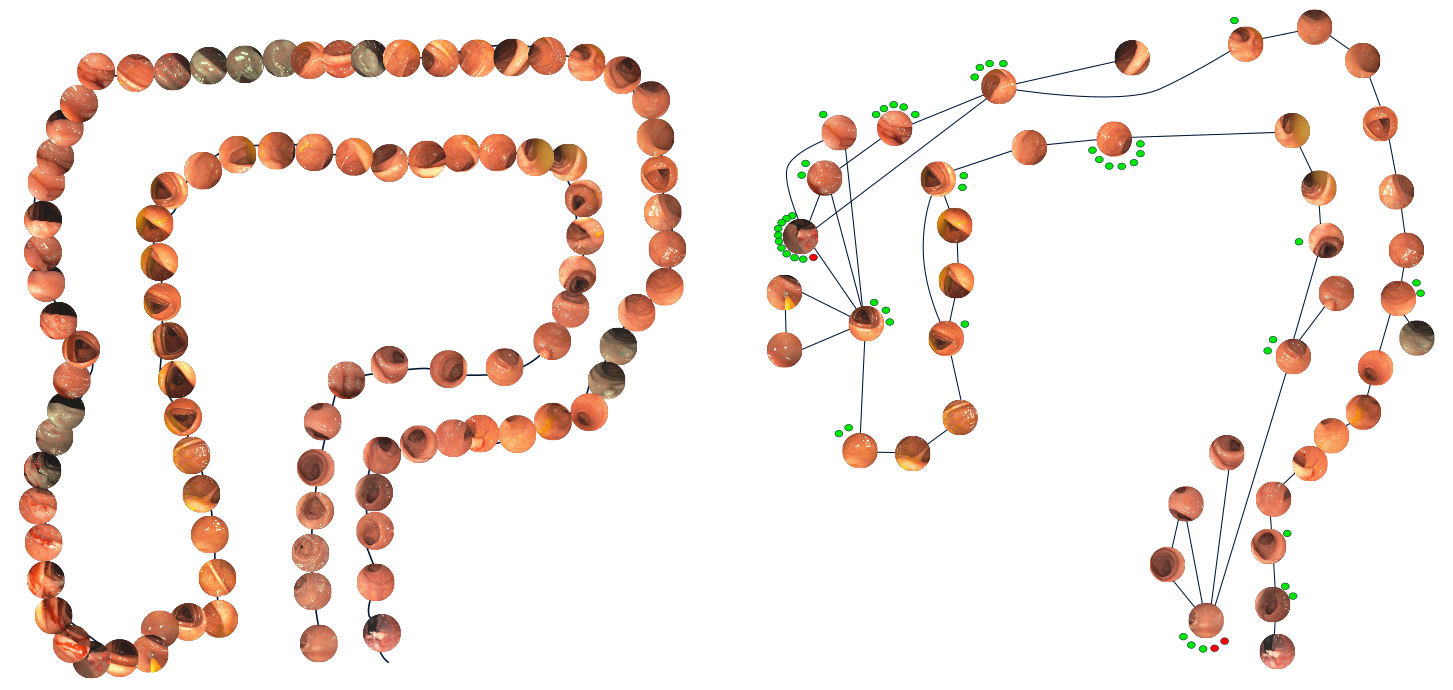}
    \caption{Seq\_027 topological map. CudaSIFT-SLAM (left) and ColonSLAM (right). }
    \label{fig:qualitative}
\end{figure}

In Figure \ref{fig:qualitative} we show a comparison between the CudaSIFT-SLAM graph and the result of our approach. Green and red dots represent correctly and wrongly localized submaps within a node, respectively. As it can be seen, we are able to build a complex graph with dozens of submaps correctly localized. The traversability connections faithfully show how the exploration was made: quickly during the entrance until the cecum was reached, showed with few traversability links, and then some exploration and back and forth movements, represented as a lot of traversability edges in the ascending colon. 

\section{Conclusions}

We have presented ColonSLAM, the first topological SLAM able to build rich graphs of the whole colon, capturing the complexity of the colonoscopy exploration. Leveraging on our robust localization network and guided by topological priors, ColonSLAM is able to reliably build a graph by finding traversability and covisibility connections between distant nodes. The graphs obtained with ColonSLAM will serve as personalized patient maps, paving the way to assisted navigation and disease monitoring in colonoscopy. In future work, we will focus on finding even longer term relationships i.e. entry-withdrawal and second explorations of the same patient as they are a limitation for ColonSLAM. Finding these long-term correspondences is the key to the building and exploitation of personalized patient maps.

\begin{credits}
\subsubsection{\ackname} Work supported by EU-H2020 grant 863146: ENDOMAPPER, Spanish grant PID2021-127685NB-I00, Arag\'on grant DGA\_T45-17R.

\subsubsection{\discintname}
The authors have no competing interests to declare that are
relevant to the content of this article.
\end{credits}

\bibliographystyle{splncs04}
\bibliography{Paper-1110}


    

    

%




\end{document}